%% file: 0.0_Main.tex
\documentclass[final,conference]{IEEEtran}
\IEEEoverridecommandlockouts
\usepackage{cite}
\usepackage{amsmath,amssymb,amsfonts}
\usepackage{algorithmic}
\usepackage{graphicx}
\usepackage{textcomp}
\usepackage{xcolor}

\usepackage{booktabs} 
\usepackage{graphicx} 
\usepackage{array}    
\usepackage{tabularx}  
\usepackage{amsmath}  
\usepackage{color}
\usepackage{soul}
\usepackage{enumitem}
\usepackage{url}

\usepackage{adjustbox} 
\usepackage{epstopdf}  
\def\BibTeX{{\rm B\kern-.05em{\sc i\kern-.025em b}\kern-.08em
    T\kern-.1667em\lower.7ex\hbox{E}\kern-.125emX}}
\begin{document}

\title{From Patient Consultations to Graphs: Leveraging LLMs for Patient Journey Knowledge Graph Construction}

\author{
    \IEEEauthorblockN{
        Hassan S. Al Khatib\textsuperscript{1},  
        Sudip Mittal\textsuperscript{1}, 
        Shahram Rahimi\textsuperscript{2},
        Nina Marhamati\textsuperscript{3},
        Sean Bozorgzad\textsuperscript{3}
    }
    \IEEEauthorblockA{
        \textsuperscript{1}Mississippi State University, Starkville, MS, USA \\
        \textsuperscript{2}University of Alabama, Tuscaloosa, AL, USA \\
        \textsuperscript{3}Potentia Analytics Inc, IL, USA \\
        Emails: hsa78@msstate.edu, mittal@cse.msstate.edu, srahimi1@ua.edu, \\ nina@potentiaco.com, sean@potentiaco.com
    }
}

\maketitle

\begin{abstract}
The transition towards patient-centric healthcare necessitates a comprehensive understanding of patient journeys, which encompass all healthcare experiences and interactions across the care spectrum. Existing healthcare data systems are often fragmented and lack a holistic representation of patient trajectories, creating challenges for coordinated care and personalized interventions. Patient Journey Knowledge Graphs (PJKGs) represent a novel approach to addressing the challenge of fragmented healthcare data by integrating diverse patient information into a unified, structured representation. This paper presents a methodology for constructing PJKGs using Large Language Models (LLMs) to process and structure both formal clinical documentation and unstructured patient-provider conversations. These graphs encapsulate temporal and causal relationships among clinical encounters, diagnoses, treatments, and outcomes, enabling advanced temporal reasoning and personalized care insights. The research evaluates four different LLMs, such as Claude 3.5, Mistral, Llama 3.1, and Chatgpt4o, in their ability to generate accurate and computationally efficient knowledge graphs. Results demonstrate that while all models achieved perfect structural compliance, they exhibited variations in medical entity processing and computational efficiency. The paper concludes by identifying key challenges and future research directions. This work contributes to advancing patient-centric healthcare through the development of comprehensive, actionable knowledge graphs that support improved care coordination and outcome prediction.

\end{abstract}


\section{Introduction}
\input{1.0_Introduction} 

\section{Background and Literature Review}
\input{2.0_Related_Work}

\section{Methodology}
\input{3.0_Methodology}

\section{PJKG Evaluation}
\input{4.0_Evaluation}

\section{Discussion}
\input{5.0_Discussion}

\section{Conclusion \& Future Work}
\input{6.0_Conclusion}

\section*{Acknowledgment}

This work was supported by NIH Grant number: R41NR02108 and the PATENT Lab (Predictive Analytics and TEchnology iNTegration Laboratory) at the Department of Computer Science and Engineering, Mississippi State University.


\end{document}

%% file: 1.0_Introduction.tex
Understanding and optimizing a patient's healthcare journey is essential to transitioning toward patient-centric healthcare. A patient journey encompasses all healthcare experiences and interactions, from the initial symptoms or diagnosis through treatment and follow-up care \cite{davies}. Despite its significance, patient data is often fragmented across multiple platforms and lacks a long-term, comprehensive representation \cite{yogesh}. This fragmentation creates significant challenges for healthcare providers in delivering coordinated care and for patients in effectively managing their health \cite{stange}. This limitation emphasizes the need for advanced frameworks to consolidate and structure disparate data sources into meaningful, actionable insights.

In this context, PJKGs emerge as an innovative solution, offering a unified and holistic representation of patient encounters across the healthcare spectrum. Unlike traditional data models that emphasize provider-centric perspectives (such as episode-based care records and discrete clinical documentation), PJKGs prioritize patient-centeredness by integrating diverse elements such as clinical records, lifestyle factors, and social determinants of health. This integration enables pattern identification and outcome predictions. Furthermore, it facilitates more effective intervention personalization. This approach ensures a more comprehensive understanding of the factors influencing health outcomes.

The motivation for developing PJKGs stems from several pressing challenges in healthcare data management and utilization. First, existing healthcare data models often operate in silos, making it difficult to trace a patient's complete medical history across different providers and systems \cite{peiris}. Second, traditional models often fail to account for the temporal, causal, and sequential relationships important for understanding patient trajectories. Third, integrating unconventional data sources, such as patient-provider encounter conversations, remains largely unexplored, leaving significant gaps in delivering genuinely personalized care.

This research addresses these challenges by leveraging a healthcare journey ontology \cite{pjo} while creating robust methods for integrating structured and unstructured healthcare data and establishing frameworks for temporal and causal relationship extraction in patient trajectories. Our research leverages and evaluates different LLMs to achieve these objectives in constructing PJKGs. LLMs offer unique capabilities in processing and understanding unstructured medical text, enabling the extraction of complex relationships and medical concepts from diverse data sources \cite{kim2024}. These knowledge graphs (KGs) are designed to map the temporal and causal relationships between clinical encounters, diagnoses, treatments, and outcomes, providing a dynamic and interconnected view of patient health trajectories. Our approach contributes to the field by introducing a novel framework for automated PJKG construction using LLMs that allows for temporal reasoning in healthcare journey mapping by integrating unstructured clinical conversations into structured knowledge graphs.

Building upon these foundations and methodological innovations, we present the organization of our research as follows: Section 2 overviews existing healthcare KGs, focusing on the methodologies and applications highlighting various patient-centric approaches. Section 3 describes the proposed methodology using different LLMs, including ontology structure, data sources, and the PJKG building process. Section 4 discusses the evaluation of PJKGs through quantitative analyses. Section 5 presents a detailed discussion of the findings. Finally, Section 6 concludes the paper by summarizing the findings and outlining directions for future research.

%% file: 2.0_Related_Work.tex
This section explores how healthcare KGs have evolved from disease-centric models to patient-centric knowledge graphs (PCKGs), enabling more individualized and data-driven care. We also highlight PCKGs' transformative potential in personalized treatment, clinical trial optimization, and disease prediction.

\subsection{Advancements in Healthcare KGs}

The evolution of KGs in healthcare signifies a paradigm
shift from traditional data systems to more complex and
interconnected networks. Early implementations of KGs, such as those rooted in electronic health records (EHRs), primarily focused on disease-centric models that integrated multimodal data for clinical problem-solving \cite{chandak, lin2023}. These graphs helped explore relationships between diseases, symptoms, and treatments, though they often lacked the granularity required for patient-centric care \cite{chen2019}.

Recent advancements have introduced PCKGs \cite{pckg}, prioritizing the holistic integration of an individual's medical history, genetic information, lifestyle behaviors, and environmental influences. PCKGs leverage structured and unstructured data sources, including clinical notes and diagnostic results, to create comprehensive patient profiles. The adoption of advanced machine learning techniques, such as named entity recognition (NER) and relationship extraction (RE), has enhanced the ability of KGs to map complex relationships within healthcare data \cite{Li2019, ruan}.

PCKGs emphasize the temporal and causal relationships
inherent in patients' data, enabling dynamic representation of health trajectories. This evolution, supported by reasoning mechanisms and semantic search techniques, allows healthcare providers to derive actionable insights for clinical decision-making \cite{lytras}. The emphasis on patient-specific data marks a critical advancement in KG technology, aligning with the goals of personalized and precision medicine.

\subsection{Applications of PCKGs}

PCKGs will transform healthcare by facilitating various applications that enhance patient care, improve diagnostic accuracy, and enable predictive modeling. One of the most significant applications is the ability to personalize treatment plans by integrating data from diverse sources to tailor interventions for individual patients \cite{gyrard}. PCKGs have been used to develop personalized treatment regimens and predict adverse drug reactions based on a patient's medical history and genetic profile. By analyzing relationships between genes, diseases, and treatments, PCKGs facilitate the identification of novel therapeutic targets and treatment approaches \cite{ping2017}.

Another application is disease prediction, where PCKGs leverage advanced analytics to identify potential health risks before symptoms manifest. Methods such as graph neural networks (GNNs) and multi-hop reasoning enable the identification of complex patterns that traditional models might overlook \cite{Sun2020}. These predictive capabilities support early intervention strategies and improve long-term health outcomes by proactively addressing conditions.

PCKGs also play an essential role in clinical trials. They streamline patient recruitment by matching individuals to trials based on detailed eligibility criteria derived from their health data. This application has been shown to increase trial efficiency and ensure better outcomes by enrolling patients whose profiles closely align with trial objectives \cite{Wang2018}.

These selected applications showcase PCKGs' transformative potential in advancing healthcare and addressing complex medical challenges.

%% file: 3.0_Methodology.tex
The construction of PJKGs requires a systematic approach that integrates diverse healthcare data, ensures semantic consistency, and addresses the complexities of patient-centered knowledge representation. This section outlines the key components of our methodology, including ontology development, data sourcing and integration, and the automated pipeline for the PJKG building process leveraging LLMs. Together, these elements form a robust framework for developing comprehensive and dynamic PCKGs.

\subsection{Ontology for PJKG}

The PJKG is supported by an ontology that captures the complexities of patient interactions throughout the healthcare continuum \cite{pjo}. By integrating structured classes, subclasses, properties, and relationships, the ontology provides a framework to represent patient journeys comprehensively, ensuring semantic consistency and enabling actionable insights.

The ontology encompasses three primary classes: \textbf{Encounter}, \textbf{IntakeForm}, and \textbf{Patient}. The \textbf{Encounter} class represents individual medical events and contains several key subclasses. The {\small\texttt{Diagnosis}} subclass captures diagnostic information with properties like diagnosis name and ICD-10 codes, while the {\small\texttt{Symptom}} subclass records patient-reported symptoms with properties including symptom name and severity. The {\small\texttt{Medication}} subclass details prescribed treatments with medication names and dosage information. The {\small\texttt{VitalSign}} subclass tracks clinical measurements including blood pressure, weight, and heart rate. The {\small\texttt{DiagTest}} subclass documents diagnostic procedures with test names and results, while the {\small\texttt{CarePlan}} subclass outlines treatment recommendations, including lifestyle changes, therapy, and follow-up actions. Finally, the {\small\texttt{Assessment}} subclass summarizes the encounter findings.

The \textbf{IntakeForm} class is divided into two main subclasses. The {\small\texttt{MedicalHistory}} subclass records past medical events, chronic conditions, medications, allergies, and family medical history. The {\small\texttt{SocialHistory}} subclass captures lifestyle factors through properties such as smoking, drinking, diet, and occupation. The \textbf{Patient} class includes properties like patient information, demographics, and insurance information. Table \ref{table:nodes} lists the nodes and the properties associated with the classes and subclasses.

Relationships between these classes establish an interconnected structure that models both static and dynamic aspects of healthcare data. Core relationships include {\small\texttt{has\_MedicalHistory}} and {\small\texttt{has\_SocialHistory}}, linking intake forms to their respective components, while others like {\small\texttt{has\_Symptom}}, and {\small\texttt{has\_Diagnosis}} connect encounters to their clinical details. Additionally, temporal and causal relationships ({\small\texttt{has\_Followup, NEXT}}, and {\small\texttt{causedBy}}) trace the sequence and dependencies of medical events, enabling a narrative representation of the patient's journey. Table \ref{table:relationships} provides a comprehensive list of these relationships and their definitions.

\begin{figure*}[h]
  \centering
  \includegraphics[draft=false, width=0.9\linewidth, keepaspectratio=true]{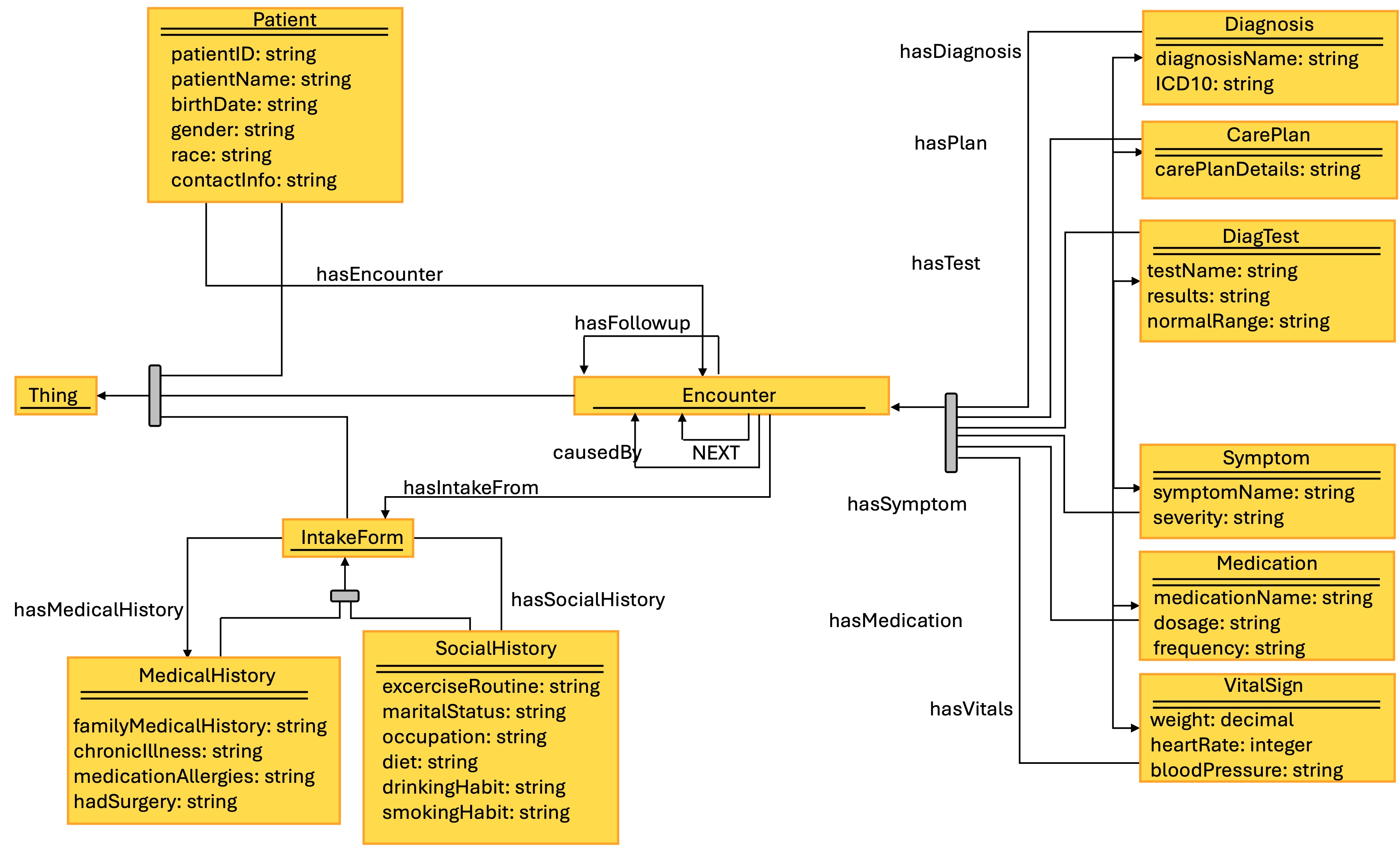}
  \caption{A visual representation of the ontology structure for the PJKG, showcasing key entities (e.g., Patient, Encounter, Diagnosis, Symptoms) and their relationships, capturing the comprehensive flow of patient care data.}
  \label{fig:ontology}
\end{figure*}

This ontology, as depicted in Fig. \ref{fig:ontology}, facilitates the integration of diverse healthcare data sources and supports advanced analytics. Structuring patient data into semantically rich entities and relationships ensures comprehensive and dynamic representation, essential for enhancing clinical decision-making and advancing patient-centric care. Ontology evaluation by Subject Matter Experts (SMEs) demonstrated its reliability and practical applicability across diverse clinical scenarios, with a Fleiss’ Kappa of 0.6117 \cite{pjo}. However, continuous refinement based on broader patient populations and use cases remains an ongoing process.

\begin{table}[htbp]
\centering
\footnotesize 
\caption{Nodes in the Knowledge Graph}
\label{table:nodes}
\begin{tabularx}{\columnwidth}{@{}lX@{}}
\toprule
\textbf{Node Name} & \textbf{Properties} \\
\midrule
Patient 
& ID, Name, DoB, Gender, Race, Contact Info, Insurance Name, Insurance ID \\
Social History 
& Exercise, Diet, Drinking, Smoking, Occupation, Marital Status, Education Level, Annual Income \\
Medical History
& Family History, Surgeries, Chronic Illnesses, Allergies, Current Medications \\
Encounter 
& Encounter Number, Date, Time \\
Diagnosis 
& Name, ICD-10 \\
Symptoms
& Name, Severity \\
Medications
& Name, Dosage \\
Diagnostic Tests
& Test Name, Results \\
Care Plan
& \textquotesingle Text\textquotesingle: Describes the next steps to be taken based on the diagnosis \\
Assessment
& \textquotesingle Text\textquotesingle: A summary of the findings by the healthcare provider \\
Vital Signs
& Blood Pressure, Heart Rate, Weight \\
\bottomrule
\end{tabularx}
\end{table}

\begin{table}[htbp]
\centering
\footnotesize
\caption{Relationships in the Knowledge Graph}
\label{table:relationships}
\begin{tabularx}{\columnwidth}{@{}lX@{}}
\toprule
\textbf{Relationship} & \textbf{Type} \\
\midrule
has\_Followup 
& For sequential encounters to follow up on a medical condition \\
NEXT          
& New encounter which is not related to the previous encounter \\
causedBy      
& If the new encounter is a referral caused by the previous encounter \\
has\_Start    
& Indicates the start of the patient journey; only between the Patient and the first Encounter \\
\bottomrule
\end{tabularx}
\end{table}

\subsection{Data Description}

This study utilized a dataset comprising six patients, evenly distributed between cardiology (n=3) and oncology (n=3) specialties, with a total of 30 clinical encounters. The dataset size reflects the inherent difficulty in acquiring real-world recorded patient-provider conversations, as such data is often subject to strict privacy regulations and requires significant anonymization. Given these constraints, this dataset is an initial proof-of-concept (POC) for constructing PJKGs using LLMs. The patient cohort included three males and three females, aged 45 to 69 years (mean age = 57.7 years). Cardiology patients (mean age = 61.7 years) presented with conditions including hypertension, atrial fibrillation, and COPD, while oncology patients (mean age = 53.7 years) primarily consisted of breast cancer cases in various treatment stages. Notable chronic conditions included hypertension (n=5, 83.3\%) and diabetes (n=2, 33.3\%), with three former smokers and three reporting no smoking history.

The dataset encompasses transcribed patient-physician conversations, with each patient participating in 4-6 encounters (mean = 5 encounters). Oncology consultations averaged 35 minutes compared to 25 minutes for cardiology visits, reflecting the complexity of cancer treatment discussions. Each patient record includes intake information documenting demographic data, medical history, current medications, and lifestyle factors. Medication profiles varied by specialty, with cardiology patients receiving cardiovascular medications (e.g., metoprolol, lisinopril) and oncology patients undergoing targeted therapies (e.g., Tamoxifen, Herceptin).

Patients' intake data included documentation of physical activity levels, ranging from limited post-surgical exercise to regular engagement in activities such as running and swimming. All data underwent thorough anonymization, with identifying information replaced by generic identifiers while maintaining clinical relevance. The conversational data captured diverse clinical scenarios, including diagnostic consultations, treatment planning, and follow-up visits.

\subsection{PJKG Construction Pipeline}

The PJKG construction process involves a systematic pipeline that transforms structured patient data and unstructured clinical encounters into a semantically rich KG. While traditional non-LLM methods such as rule-based entity extraction could be used for NER and RE; these approaches require extensive domain-specific feature engineering and struggle with implicit relationships present in conversational data. Our primary focus was to evaluate whether general-purpose LLMs could perform these tasks without requiring extensive manual rule-setting. In this section, we detail our methodology for constructing PJKGs through a multi-stage process that combines structured data integration, LLM-based information extraction from unstructured encounter data, quality control, and graph database operations. Different stages of this process are described in the following subsections: 

\subsubsection{Patient Profile Construction}
The foundation of each PJKG begins with creating a comprehensive patient profile node in Neo4j. Let $P = \{p_1, p_2, ..., p_n\}$ represent the set of patients, where each patient $p_i$ has associated profile data $\mathcal{D}_i$ comprising demographic information, medical history $\mathcal{M}_i$, and social history $\mathcal{S}_i$. The profile construction process establishes the primary patient node and its immediate relationships:

\[
p_i \xrightarrow{\text{HAS\_PROFILE\_DATA}} \mathcal{D}_i
\]
\[
p_i \xrightarrow{\text{HAS\_MEDICAL\_HISTORY}} \mathcal{M}_i
\]
\[
p_i \xrightarrow{\text{HAS\_SOCIAL\_HISTORY}} \mathcal{S}_i
\]

\subsubsection{LLM-based Entity and Relationship Extraction}
Following patient profile establishment, we process clinical encounter transcripts $\mathcal{E}_i = \{e_{i1}, e_{i2}, ..., e_{in}\}$ for each patient $p_i$, where $n$ represents the number of encounters. The extraction process utilizes a carefully engineered prompt-based approach.

\subsubsection{Prompt Engineering for Structured Extraction}
The prompt template is designed to ensure consistent and valid JSON output structured according to our ontology. Key components of the prompt include:

\begin{itemize}
    \item Explicit structural requirements for JSON validity
    \item Pre-defined entity categories aligned with the ontology
    \item Relationship type constraints (e.g., \small{\texttt{has\_followup}}, \small{\texttt{causedby}}, \small{\texttt{NEXT}})
    \item Default value specifications for handling missing information
\end{itemize}

For each encounter $e_{ij}$, we generate a prompt $\mathcal{P}_{ij}$ that includes:

\[
\mathcal{P}_{ij} = \{\text{encounter\_id}, \text{date}, \text{time}, \text{transcript\_text}\}
\]

\subsubsection{Processing Pipeline}
The extraction process follows a defined workflow:

\[
\mathcal{J}_{ij} = \text{LLM}(\mathcal{P}_{ij}, e_{ij})
\]
where $\mathcal{J}_{ij}$ represents the structured JSON output containing extracted entities and relationships for encounter $e_{ij}$.

\subsubsection{Quality Control and Validation}
The validation process ensures data integrity through multiple checks:

\begin{enumerate}[label=\alph*)]
    \item \textbf{Syntactic Validation}: Verifies JSON structure adherence:
    \[
        \text{Valid}(\mathcal{J}_{ij}) = \begin{cases} 
            \text{True}, & \text{if } \mathcal{J}_{ij} \text{ follows schema } \mathcal{S} \\
            \text{False}, & \text{otherwise}
        \end{cases}
    \]
    
    \item \textbf{Semantic Validation}: Ensures entity and relationship consistency with the ontology:
    \[
        E_{ij} \subseteq \mathcal{O}_E, \quad R_{ij} \subseteq \mathcal{O}_R
    \]
    where $\mathcal{O}_E$ and $\mathcal{O}_R$ represent ontology-defined entities and relationships, respectively.
    
    \item \textbf{Temporal Consistency}: Validates encounter sequence integrity:
    \[
        \forall e_{ij}, e_{ik}: j < k \implies \text{timestamp}(e_{ij}) < \text{timestamp}(e_{ik})
    \]
\end{enumerate}

\subsubsection{Knowledge Graph Integration}
The final stage involves constructing the PJKG in Neo4j. For each validated JSON output $\mathcal{J}_{ij}$, we create a subgraph:

\[
    \mathcal{G}_{ij} = (V_{ij}, E_{ij})
\]
where $V_{ij}$ represents the nodes (entities) and $E_{ij}$ represents the edges (relationships). The complete PJKG for the patient $p_i$ is constructed as:

\[
    \text{PJKG}_i = \bigcup_{j=1}^n \mathcal{G}_{ij}
\]

The integration process maintains referential integrity through unique identifiers and enforces relationship constraints defined in the ontology. Each encounter node is linked to its corresponding patient through:

\[
    p_i \xrightarrow{\text{HAS\_START}} e_{ij}
\]

\begin{figure*}[h]
  \centering
  \includegraphics[draft=false, width=0.9\linewidth, keepaspectratio=true]{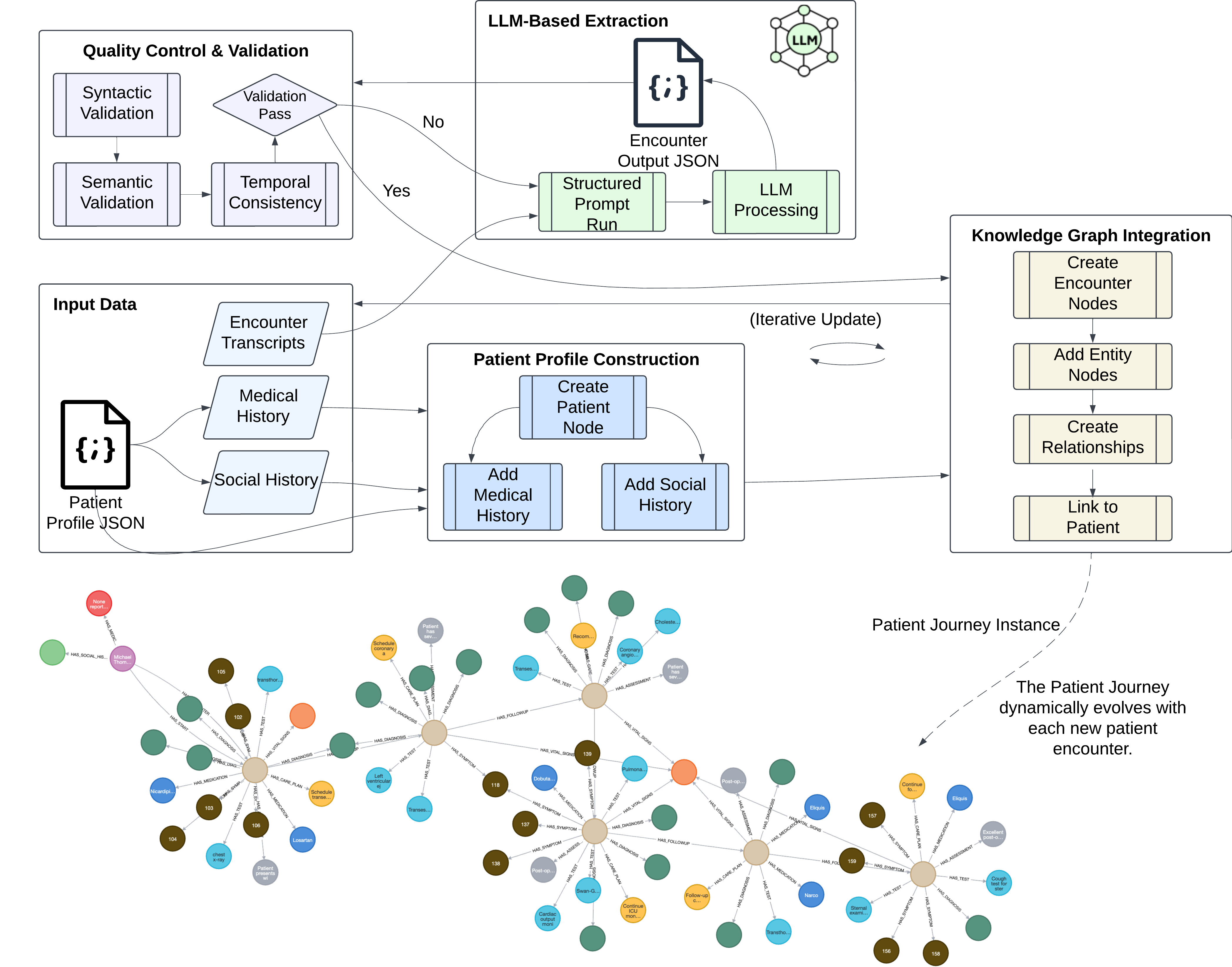}
  \caption{A visual depiction of the process to create a PJKG using ontology-based structured prompts and transcribed patient-provider conversations, processed by an LLM to extract NER and RE into JSON format, then loaded into graph DB for visualization and analysis.}
  \label{fig:process_v2}
\end{figure*}

This methodological approach, as depicted in Figure \ref{fig:process_v2}, ensures the creation of semantically rich, temporally ordered, and structurally consistent PJKGs that capture the complexity of clinical narratives while maintaining data quality and ontological alignment.

\subsection{Mapping the Journey: A Use Case of Evolving PJKG}

\begin{figure*}[h]
  \centering
  \includegraphics[draft=false, width=0.9\linewidth, keepaspectratio=true]{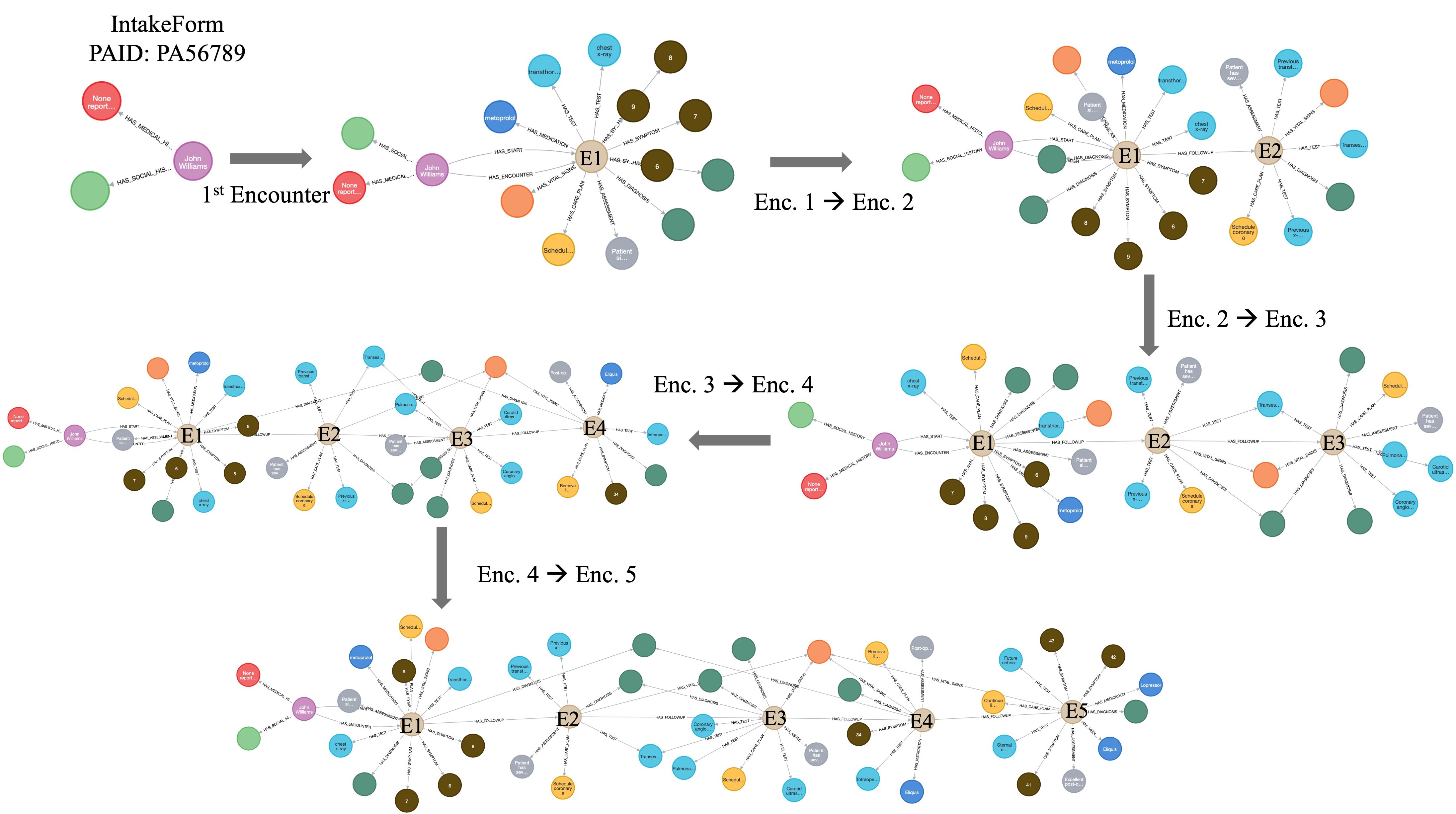}
  \caption{Visualizing the dynamic evolution of a PJKG across five clinical encounters, highlighting the integration of new diagnostic, treatment, and care plan information at each stage of the patient’s journey.}
  \label{fig:growth}
\end{figure*}

The PJKG evolves dynamically by capturing each clinical encounter into an interconnected structure, as demonstrated in Figure \ref{fig:growth} through the journey of a 64-year-old male patient. The initial intake established the foundation by recording demographic details, social history, and prior medical conditions, including controlled hypertension managed with metoprolol. When the patient presented with symptoms of shortness of breath, dizziness, and reduced exercise tolerance, these formed the first nodes in the knowledge graph.

The patient's diagnostic journey unfolded through two critical encounters. The first encounter (E1) confirmed mitral valve regurgitation (ICD-10: I34.0) through transthoracic echocardiography, supported by chest X-ray findings. In the follow-up encounter (E2), a transesophageal echocardiogram refined the diagnosis to severe mitral valve regurgitation with preserved ejection fraction, leading to referrals for coronary angiography and surgical evaluation.
The treatment phase progressed through subsequent encounters. The preoperative encounter (E3), caused by (E2), confirmed the severity of the mitral valve condition and established the surgical plan. The postoperative phase (E4), a follow-up from (E3), documented the successful repair with an annuloplasty ring and no residual regurgitation, incorporating nodes for intraoperative echocardiograms and medications. By the fifth encounter (E5), the graph captured the recovery trajectory, documenting resolved symptoms, sternal stability, and long-term care plans.

Through this sequence, the PJKG demonstrates its capability to capture a comprehensive and evolving view of the patient's clinical journey. By linking initial symptoms to diagnostic insights, interventions, and recovery milestones, clinicians can contextualize decisions within the broader narrative of the patient's health, fostering improved continuity of care.

%% file: 4.0_Evaluation.tex
This section provides a comprehensive analysis of the
quality and performance of PJKGs generated by four
LLMs: Athropic (claude-3-5-sonnet-latest), Mistral (mistral-large-2411), Llama 3.1, and OpenAI (chatgpt-4o-latest). The evaluation encompasses structural metrics (Instantiated Class Ratio - ICR, Instantiated Property Ratio - IPR, and graph completeness), computational metrics (latency, throughput, and scalability), and semantic accuracy assessment through comparison
with expert-annotated ground truth data. The semantic evaluation examines both entity and relationship extraction performance using precision, recall, and F1 scores. While entity extraction results were generally strong across models, relationship extraction posed a greater challenge. This difficulty arises because clinical relationships often require contextual understanding, temporal reasoning, and implicit knowledge, which current general-purpose LLMs may not fully capture. Additionally, conversational data frequently contains fragmented or ambiguous references that complicate relationship annotation. This approach offers a holistic understanding of each LLM's strengths and limitations in generating structurally sound, computationally efficient, and clinically accurate KGs, providing insights for improving real-world applications.

\subsection*{Structural Metrics}
The structural evaluation focuses on assessing schema compliance of LLM-generated KGs through the Instantiated Class Ratio (ICR), Instantiated Property Ratio (IPR), and graph completeness metrics. These metrics quantify the alignment of instantiated elements with schema-defined classes and properties.

The \textbf{Instantiated Class Ratio (ICR)} measures the proportion of schema-defined classes instantiated in the KG \cite{seo2022}, calculated as:
\[
\text{ICR} = \frac{C_{\text{instantiated}}}{C_{\text{total}}}
\]
where \( C_{\text{instantiated}} \) is the number of instantiated classes, and \( C_{\text{total}} \) is the total number of schema-defined classes. All models achieved an ICR of 1.00, indicating full instantiation of schema-defined classes.

Similarly, the \textbf{Instantiated Property Ratio (IPR)} evaluates the instantiation of schema-defined properties \cite{seo2022}, using the formula:
\[
\text{IPR} = \frac{P_{\text{instantiated}}}{P_{\text{total}}}
\]
where \( P_{\text{instantiated}} \) is the number of instantiated properties, and \( P_{\text{total}} \) is the total schema-defined properties. Like ICR, all models achieved an IPR of 1.00, reflecting robust schema compliance.

To extend the structural analysis, \textbf{Graph Completeness Metrics} evaluate the completeness of nodes, relationships, and properties \cite{zhang2024}:

\begin{small}
\[
\begin{array}{l}
\text{Node Completeness} = \frac{\text{No. of Instantiated Nodes}}{\text{Total Required Nodes in Schema}} \times 100, \\[1.5ex]
\text{Relationship Completeness} = \frac{\text{No. of Instantiated Relationships}}{\text{Total Required Relationships in Schema}} \times 100, \\[1.5ex]
\text{Property Completeness} = \frac{\text{No. of Instantiated Properties}}{\text{Total Required Properties in Schema}} \times 100.
\end{array}
\]
\end{small}

\begin{figure}[h]
  \centering
  \includegraphics[draft=false, width=0.95\linewidth, keepaspectratio=true]{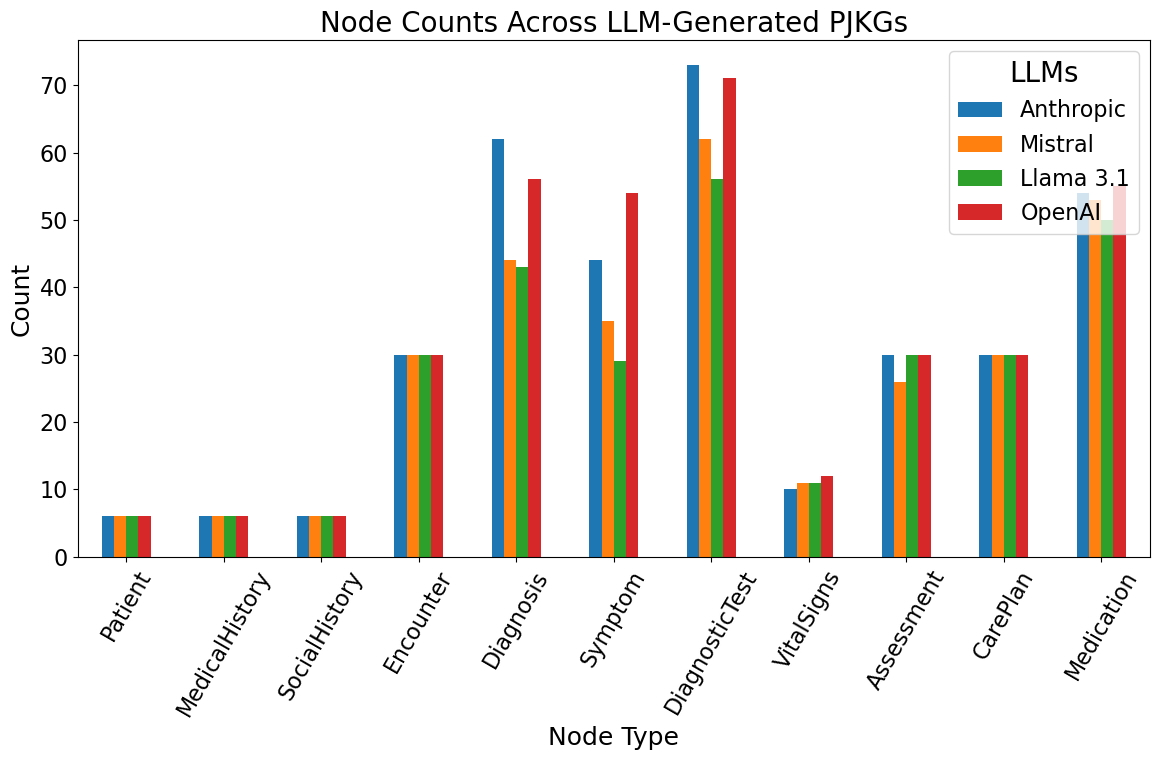}
  \caption{Comparison of node counts across PJKGs generated by different LLMs. The chart illustrates the distribution of node types, such as \texttt{Symptom}, \texttt{Diagnosis}, and \texttt{DiagnosticTest}, reflecting differences in node completeness across LLM-generated graphs.}
  \label{fig:n_complete}
\end{figure}

\begin{figure}[h]
  \centering
  \includegraphics[draft=false, width=0.95\linewidth, keepaspectratio=true]{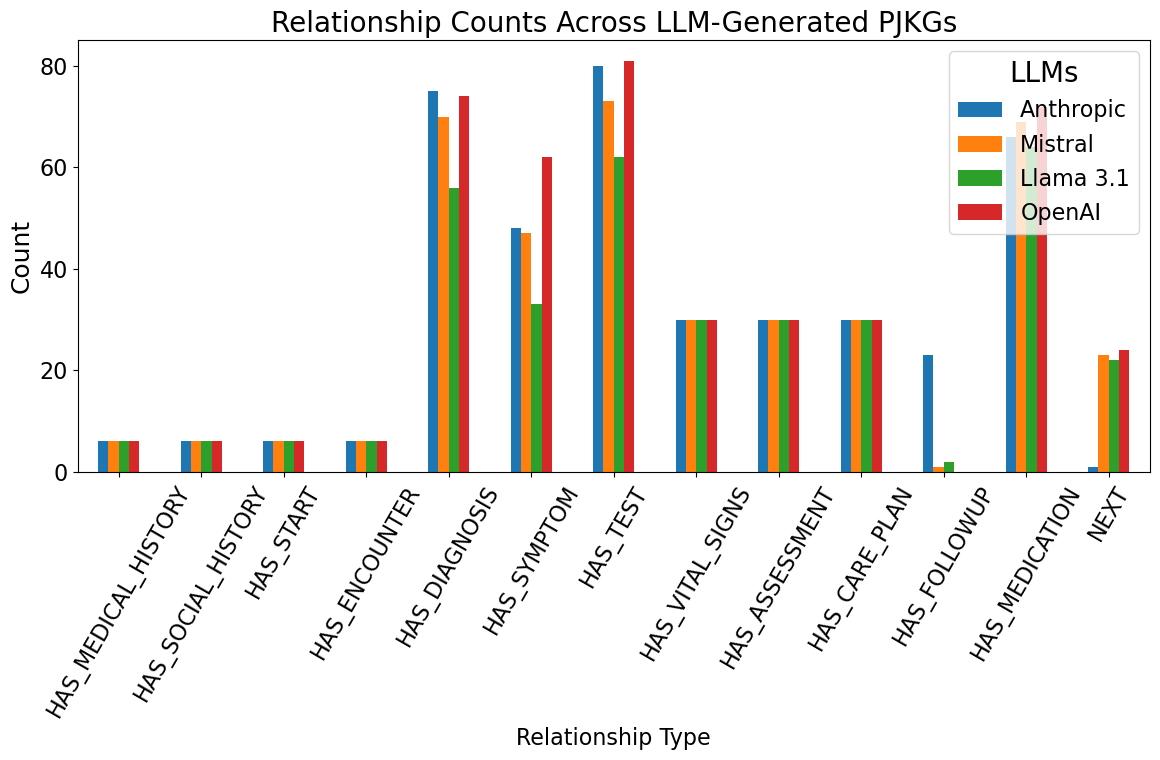}
  \caption{Comparison of the relationship counts across PJKGs built using different LLMs. The chart highlights the distribution of various relationship types, such as \texttt{HAS\_DIAGNOSIS}, \texttt{HAS\_SYMPTOM}, and \texttt{HAS\_TEST}, indicating variations in the completeness of relationship representation.}
  \label{fig:r_complete}
\end{figure}

All PJKGs achieved 100\% completeness in nodes, relationships, and properties (Figures \ref{fig:n_complete}, and \ref{fig:r_complete}), demonstrating their structural adherence to the ontology (Figure 1).

Despite perfect ICR, IPR, and graph completeness scores, variations in node and relationship counts were observed. Anthropic generated higher counts for \texttt{Diagnosis} and \texttt{DiagnosticTest} nodes and \texttt{HAS\_DIAGNOSIS} relationships, reflecting detailed processing of medical entities. In contrast, Llama 3.1 produced fewer instances in categories like \texttt{Symptom} and \texttt{Diagnosis} but achieved parity in key schema elements like \texttt{Assessment} and \texttt{Care Plan}. These differences suggest variability in how schema elements are interpreted and prioritized during graph construction. 

\subsection*{Computational Metrics}
The computational evaluation assesses the operational efficiency of Neo4j-stored KGs through latency, throughput, and scalability metrics. 

\textbf{Latency} measures the average query response time:
\[
\text{Latency} = \frac{\sum_{i=1}^{N} t_i}{N}
\]
where \( t_i \) is the time for the \( i^{\text{th}} \) query, and \( N \) is the total number of queries. Lower latency indicates faster performance. 

\textbf{Throughput} evaluates the number of queries processed per second:
\[
\text{Throughput} = \frac{Q_{\text{total}}}{T_{\text{total}}}
\]
where \( Q_{\text{total}} \) is the total number of queries, and \( T_{\text{total}} \) is the total execution time. Higher throughput reflects better scalability.

\begin{figure*}[h]
  \centering
  \includegraphics[draft=false, width=0.9\linewidth, keepaspectratio=true]{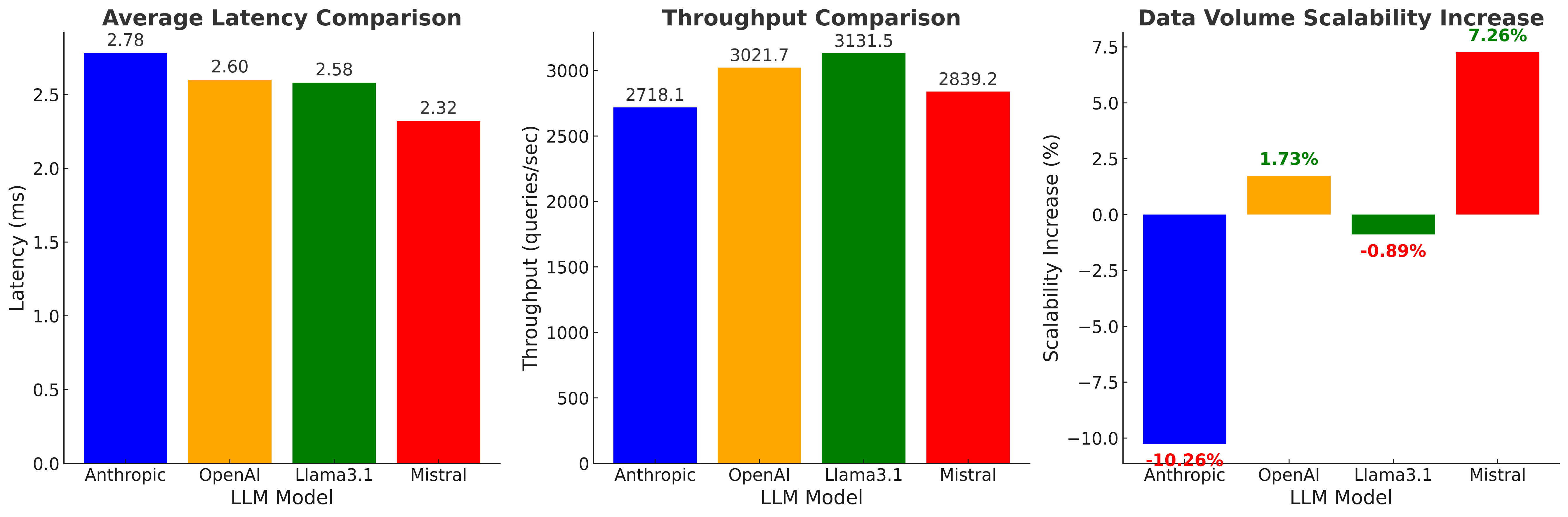}
  \caption{Performance comparison of LLM-generated PJKGs across three metrics: (a) Average latency in milliseconds; (b) Throughput in queries per second; and (c) Data volume scalability increases as a percentage}
  \label{fig:computation}
\end{figure*}

The comparative analysis illustrated in Figure \ref{fig:computation} reveals diverse performance patterns across the LLM-generated knowledge graphs. Mistral demonstrates superior performance in two key metrics: achieving the lowest latency (2.32ms) and the best scalability with a 7.26\% increase in data volume handling capability. While Llama3.1 achieves the highest throughput (3131.50 queries/sec) and maintains competitive latency (2.58ms), it shows slightly negative scalability (-0.89\%). OpenAI presents balanced performance with moderate latency (2.60ms), strong throughput (3021.70 queries/sec), and positive scalability (1.73\%). In contrast, Anthropic faces performance challenges with the highest latency (2.78ms), lowest throughput (2718.10 queries/sec), and poorest scalability (-10.26\%), suggesting potential limitations in handling increasing data volumes.

Scalability analysis further highlighted these differences. Llama 3.1 showed minimal performance degradation, with a 0.45\% increase in response time as data volumes grew, demonstrating excellent scalability. Conversely, OpenAI faced a significant scalability challenge, with a -6.63\% change, indicating performance degradation under load. Anthropic and Mistral demonstrated moderate computational performance and scalability, providing balanced alternatives for scenarios where computational demands are less critical. Table \ref{table:performance_comparison} presents a comparative analysis of performance metrics across PJKGs built by four different LLM models, highlighting variations in latency, throughput, and data volume scalability.

\subsection*{Semantic Accuracy Assessment}

To evaluate the effectiveness of LLM-constructed PJKGs, we developed a ground truth KG based on clinical encounters with two cardiology patients (PA56789 and PM82487). The encounters were manually annotated by a medical doctor to establish accurate entity and relationship extraction benchmarks. We then compared the LLM-constructed PJKGs against this ground truth using precision, recall, and F1-score metrics for both entities and relations, allowing us to assess the accuracy of the extracted information.

Our analysis revealed performance variations among the LLM-constructed PJKGs, with Anthropic's PJKG demonstrating superior performance across all metrics (Combined F1: 0.73). In entity extraction, Anthropic's PJKG achieved the highest entity F1-score (0.73) with a precision of 0.68 and recall of 0.79, indicating strong accuracy and coverage. In relationship extraction, Anthropic's PJKG significantly outperformed other PJKGs with consistent precision and recall values of 0.73. In comparison, other PJKGs showed notably lower performance, with OpenAI's PJKG achieving the second-best entity extraction results (F1: 0.70, precision: 0.62, recall: 0.80) but struggling with relationships (F1: 0.18, precision: 0.18, recall: 0.18). The consistently lower performance in relationship extraction across Llama3, Mistral, and OpenAI PJKGs suggests that more complex approaches may be needed to improve relationship extraction capabilities in medical KG construction. Future work should focus on enhancing relationship extraction performance through specialized medical context understanding and improved entity relationship modeling.

\begin{table}[htbp]
\centering
\footnotesize
\caption{Performance Metrics Comparison Across PJKGs}
\label{table:performance_comparison}
\begin{tabularx}{\columnwidth}{@{}XXXX@{}}
\toprule
\textbf{LLM Model} & \textbf{Latency (ms)} & \textbf{Throughput (queries/sec)} & \textbf{Scalability Increase (\%)} \\
\midrule
Anthropic & 2.78 & 2,718.10 & -10.26 \\
OpenAI & 2.60 & 3,021.70 & 1.73 \\
Llama3.1 & 2.58 & 3,131.50 & -0.89 \\
Mistral & 2.32 & 2,839.20 & 7.26 \\
\bottomrule
\end{tabularx}
\end{table}

%% file: 5.0_Discussion.tex
This study investigated the effectiveness of different LLMs in constructing PJKGs that can integrate structured patient data and unstructured clinical conversations that support temporal reasoning and patients' medical journey mapping. The evaluation revealed valuable findings about the structural, semantic, and computational performance of the PJKGs generated by four different LLMs. One key consideration in our approach was the decision to use general-purpose LLMs without fine-tuning on medical-specific datasets. Fine-tuning could potentially improve NER and RE; however, it requires a large volume of high-quality labeled data, which is currently limited in this domain. Moreover, fine-tuned models may suffer from overfitting to specific clinical terminologies and lose adaptability to broader medical discourse. In our case, all PJKGs demonstrated perfect structural compliance with the schema, evidenced by an ICR and IPR of 1.00, and 100\% completeness in nodes, relationships, and attributes. Semantic accuracy assessments revealed notable performance gaps, with Anthropic achieving the highest semantic accuracy (Combined F1: 0.73), including an entity F1-score of 0.73 and a relationship F1-score of 0.73. OpenAI, while second-best in entity extraction (F1: 0.70), struggled with relationship extraction (F1: 0.18).

Statistical analysis of class instantiation revealed significant variations across LLMs in medical entity processing. Anthropic demonstrated the highest diagnostic-related instances, with 62 \texttt{Diagnosis} and 73 \texttt{DiagnosticTest} instances, exceeding Llama 3.1's counts by 44\% and 30\%, respectively. Llama 3.1 showed the lowest \texttt{Symptom} instance count at 29, compared to OpenAI's maximum of 54 instances.  
The computational evaluation revealed Llama 3.1 as the most efficient model, with its latency (0.33 ms) being 84\% lower than other models' average (2.13 ms). Its throughput (3641.50 queries/sec) exceeded OpenAI's performance by 33\%. Llama 3.1 showed minimal performance impact (0.45\% increase) as data volumes grew, while OpenAI experienced the most significant degradation (-6.63\%). Anthropic and Mistral demonstrated moderate computational performance, with throughput rates of 2983.10 and 2859.10 queries/sec, respectively.

Based on these findings, technical improvements for enhancing PJKG's performance include addressing semantic accuracy gaps through improved entity and relationship modeling, query optimization through indexing strategies, and graph compression techniques for high-volume medical data. These enhancements, combined with the demonstrated efficiency of LLM-generated PJKGs, suggest promising applications in healthcare settings where comprehensive patient journey analysis is crucial.

%% file: 6.0_Conclusion.tex
This paper presents a detailed methodology for constructing PJKGs using structured (intake forms) and unstructured (conversational) data. It demonstrates the feasibility and effectiveness of LLMs in extracting and structuring patient-centric clinical information. The proposed framework successfully transforms unstructured patient-physician dialogues and patient-structured data into semantically rich and actionable KGs. These contributions mark a step toward advancing personalized healthcare through data-driven insights.

While these results demonstrate the potential of PJKGs, several important challenges emerged during our investigation. Semantic accuracy variation, particularly in entity and relationship extraction across LLM models, highlights the need for more robust and consistent data extraction methods. Anthropic's higher semantic accuracy highlights the potential of advanced approaches, but gaps remain, particularly in balancing precision and recall for complex clinical relationships. Additionally, while the current implementation effectively captures conversational data, it lacks the integration of EHRs, which are important for creating a more comprehensive view of patient journeys. Addressing these limitations is important to enhance the utility and generalizability of PJKGs in diverse healthcare settings.

Looking ahead, our future work will prioritize several key areas. First, enhancing the scalability and diversity of datasets will be crucial to accommodate larger patient populations and a broader spectrum of medical conditions. Second, integrating EHR data with unstructured conversational insights will enable a more comprehensive representation of patient journeys. Third, advances in LLMs and ontology design should be leveraged to extract more nuanced clinical information, thus improving the precision, recall, and relevance of PJKGs. Additionally, enhancing relationship extraction performance will be a key focus, particularly through specialized medical context understanding and improved entity-relationship modeling, as accurate relationship representation is critical for generating clinically meaningful KGs.

Beyond methodological advancements, an important aspect of translating this research into clinical practice involves real-world deployment considerations. To be practically useful in healthcare settings, a PJKG system must integrate seamlessly with EHRs and hospital data management systems. Key challenges include ensuring compliance with data privacy regulations (such as HIPAA), maintaining real-time processing capabilities, and designing an interpretable and actionable interface for clinicians. Future developments could potentially involve pilot studies within clinical institutions to evaluate feasibility and adoption barriers. Finally, addressing scalability through advanced graph management techniques and data compression will ensure that large datasets are handled efficiently, supporting widespread implementation in real-world healthcare environments.